\documentclass{article}
\usepackage[preprint]{corl_2026}
\usepackage[T1]{fontenc}
\usepackage{float} 
\usepackage{amsmath,amssymb,booktabs,graphicx,microtype,tabularx}
\newcommand{\CalVLA}{\textsc{CalVLA}}
\newcommand{\ACal}{\textsc{ACal}}
\newcommand{\auc}{\mathrm{AUROC}}
\newcommand{\tsr}{\mathrm{TSR}}
\title{CalVLA: Shaping Native Confidence in Vision-Language-Action Policies with Outcome Supervision}
\author{Boniface Chukwuemeka Godwin\\
Carnegie Mellon University Africa\\
Kigali, Rwanda\\
\texttt{bgodwin@andrew.cmu.edu}}
\hypersetup{pdfauthor={Boniface Chukwuemeka Godwin},
  pdftitle={CalVLA: Shaping Native Confidence in Vision-Language-Action Policies with Outcome Supervision},
  pdfsubject={Science of Physical AI Safety (SPAIS) Workshop at CoRL 2026}}
\begin{document}
\raggedbottom
\maketitle
\begin{abstract}
Can a vision-language-action (VLA) policy signal likely task failure from its own action probabilities? We introduce \CalVLA{}, which shapes native action-token confidence using successful and failed rollouts and an Action-Confidence Alignment (\ACal{}) squared margin-ranking loss. Low-rank adapters update OpenVLA-7B without an auxiliary failure detector. Observations are paired within tasks at similar normalized elapsed times, not necessarily equivalent physical progress. On a held-out test split from 500 LIBERO-10 configurations, adapted confidence achieves a joint failure-discrimination AUROC of $0.7392\pm0.0480$ over three seeds, averaging scores at 130 and 260 steps. Frozen-policy probes reach 0.6146--0.6436, but superiority over the linear probe is statistically inconclusive. For the same confidence summary, AUROC rises from 0.4485 before adaptation to 0.7327 after adaptation. Mean task success decreases from 58\% to 55\%. These results show improved failure discrimination, not improved control or established safety benefits.
\end{abstract}
\keywords{VLA policies, Native confidence, Failure discrimination}
\vspace{-0.7em}
\begin{figure}[H]
\centering
\includegraphics[width=\linewidth]{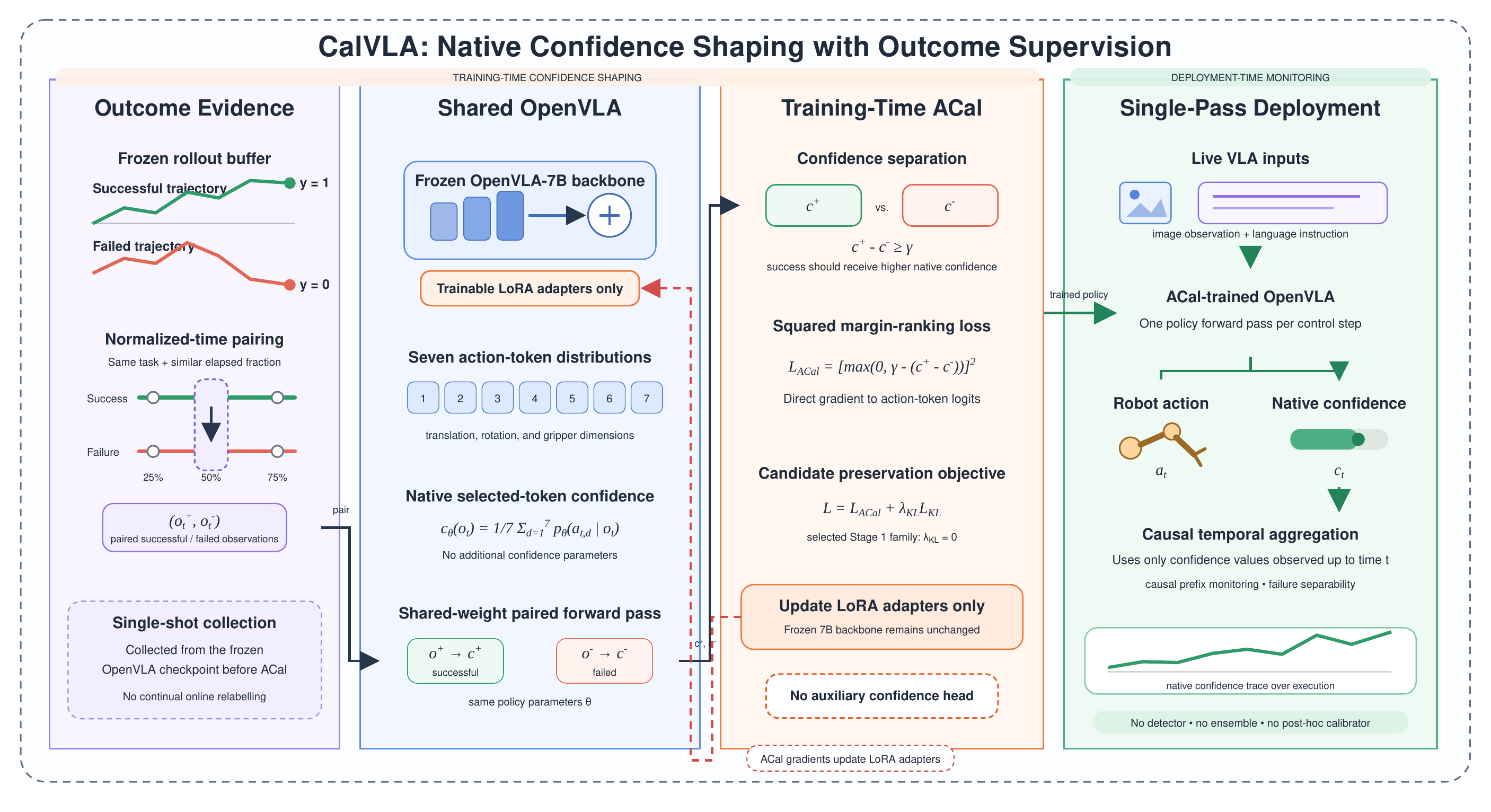}
\caption{\CalVLA{} matches successful and failed observations within the same task by normalized elapsed time, not necessarily equivalent physical task progress. \ACal{} adapts LoRA parameters using native action-token confidence. At test time, observed-prefix confidence supplies a failure-ranking score without a separately trained detector.}
\label{fig:calvla_framework}
\end{figure}

\section{Introduction and related work}
\label{sec:introduction}
A robot can keep issuing confident actions after a missed grasp or an unsuccessful placement. For vision-language-action (VLA) policies, this is a training mismatch: imitating demonstrated actions does not explicitly teach the model whether its behavior will ultimately succeed. OpenVLA~\citep{kim2024openvla} already produces probabilities over discrete action tokens. We call a summary of these probabilities \emph{native confidence}, because it comes from the action-producing policy itself.

\paragraph{Confidence and temporal information.}
Raw action probabilities need not predict task outcomes~\citep{zollo2025confidence}. Shifting Uncertainty aggregates signals across time and action dimensions~\citep{tang2026shifting}, while temporal-difference calibration learns sequential success confidence through value estimation~\citep{francismeretzki2026td}. These approaches motivate studying what confidence reveals over an unfolding attempt; our objective instead changes the native action probabilities through outcome-supervised policy adaptation.

\paragraph{Learned monitors and recovery.}
SAFE and SAFECAST learn failure predictors from policy features~\citep{gu2025safe,rajaprakash2026safecast}; Hide-and-Seek uses rollout labels~\citep{park2026hide}; and ActProbe and INSIGHT exploit action chunks or token uncertainty~\citep{huang2026actprobe,karli2026insight}. Other work studies consistency-based monitors~\citep{agia2025unpacking}, success-only examples~\citep{xu2025faildetect}, or recovery~\citep{ma2026cyclevla}. Our contribution instead \emph{modifies the policy's native action-token confidence} through outcome supervision. Frozen-feature probes isolate the detector-versus-controller choice; they are not reproductions of those prior systems.

\paragraph{\CalVLA{}} ranks confidence for successful versus failed attempts of the same task at similar \emph{normalized elapsed times}, not verified equivalent physical stages. We measure \emph{failure discrimination} (ranking eventual failures above successes) and task-success rate (TSR). Discrimination is distinct from probability calibration~\citep{guo2017calibration} and does not establish the benefit of a recovery intervention. This study tests whether the confidence signal can improve without an extra detector, and whether adapting the controller changes its task performance.

\section{Method}
\label{sec:method}
\paragraph{Native confidence.}
An observation $o$ contains the current image and instruction. OpenVLA generates seven action tokens autoregressively, one per action dimension. Write $p_{\theta,d}(k\mid o)$ for the probability of bin $k$ in dimension $d$, suppressing the preceding tokens. We train the mean top-token confidence
\begin{equation}
 c_{\mathrm{mean},\theta}(o)=\frac{1}{7}\sum_{d=1}^{7}\max_k p_{\theta,d}(k\mid o).
 \label{eq:mean_confidence}
\end{equation}
With deterministic decoding, each maximum is the probability of the selected token. The score is differentiable almost everywhere; a high value indicates concentrated action probabilities, not a known probability of task success.

\paragraph{Learning from outcomes.}
For a rollout of $L$ steps indexed $t=1,\ldots,L$, define normalized elapsed time $\rho=t/L$. We pair observations $(o^+,o^-)$ from successful and failed rollouts of the same task when $|\rho^+-\rho^-|\leq0.02$. Different durations or termination conditions mean that paired observations may represent different physical task stages. The loss is
\begin{equation}
 \mathcal L_{\mathrm{ACal}}=\left[\max\left(0,\gamma-c_{\mathrm{mean},\theta}(o^+)+c_{\mathrm{mean},\theta}(o^-)\right)\right]^2,
 \qquad\gamma=0.2.
 \label{eq:acal}
\end{equation}
This squared hinge-style ranking objective vanishes when success-associated confidence exceeds failure-associated confidence by at least the margin. Same-task, relative-time pairing reduces task-identity and timing shortcuts, but does not match physical states or identify the failed action; a failed rollout may include correct early actions.

\paragraph{Adaptation and scoring.}
We update rank-16 LoRA adapters~\citep{hu2022lora} in attention $q_{\mathrm{proj}}$ and $v_{\mathrm{proj}}$, keeping pretrained weights frozen. Validation compares candidate strengths of a forward Kullback--Leibler (KL) penalty toward the original policy and selects zero; the reported models therefore optimize Equation~\eqref{eq:acal} alone. At an evaluation step $h$, failure risk is the negative observed-prefix mean,
\begin{equation}
 r_{i,h}=-\frac{1}{h}\sum_{t=1}^{h}c_{\mathrm{mean},\theta}(o_{i,t}).
 \label{eq:runtime_risk}
\end{equation}
This can be updated online without future observations. We evaluate ranking, without selecting an alarm threshold or executing a recovery action.

\section{Experimental protocol}
\label{sec:experiments}
\paragraph{Data and training.}
We evaluate the prescribed grid of ten LIBERO-10 tasks with 50 initial states each~\citep{liu2023libero}. An outcome-blind rule fixes 350 training, 50 validation, and 100 test configurations, with 35/5/10 per task. Frozen OpenVLA succeeds on 243 of all 500 configurations and 58 test configurations. Training uses 11,378 same-task pairs, with failed observations restricted to the final 130 steps and at most 128 pairs per source episode. Each seed makes one pass using AdamW, learning rate $10^{-5}$, and batch size one pair.

We train six KL-coefficient families, $\lambda_{\mathrm{KL}}\in\{0,1.5,5,15,50,150\}$, with three seeds each. Eligibility requires mean validation task-success rate at least 0.42 and every seed at least 0.38. We select the highest mean joint AUROC among eligible families, resolving differences within 0.01 toward smaller coefficients. This selects $\lambda_{\mathrm{KL}}=0$; only its three seeds are evaluated on the locked test set. Appendices~\ref{app:split}--\ref{app:selection} give the split and training details.

\paragraph{Controls.}
The predefined frozen-native baseline uses the least confident action dimension, $c_{\mathrm{min},\theta}(o)=\min_d\max_k p_{\theta,d}(k\mid o)$. Because \ACal{} trains mean-top confidence, we also evaluate both policies with the \emph{same min-top summary}. Three outcome-supervised probes on frozen 4096-dimensional representations preserve the controller: P-LINEAR (sigmoid linear readout, 4,097 parameters), P-STATIC (width-128 MLP), and P-TEMPORAL (width-128 MLP with a trailing 25-step representation mean). These controlled probes do not reimplement prior monitoring systems.

\paragraph{Evaluation.}
AUROC ranks failures above successes (0.5 is chance). Joint AUROC averages landmark scores at 130 and 260 steps; at 260 only still-active trajectories are evaluated, so landmark populations differ. We report TSR and means and sample SD across adaptation seeds. Paired, task-stratified bootstrap intervals resample held-out configurations within task, preserving comparator pairing. They quantify resampling uncertainty on this deterministic test grid, not real-robot variability.

\section{Results and discussion}
\label{sec:results}
\begin{table}[!htbp]
\centering\small
\caption{Locked-test results. \ACal{} uses mean top-token confidence; frozen native uses min-top confidence. Joint AUROC and TSR for \ACal{} show mean $\pm$ sample SD across three seeds. Probes preserve frozen task success.}
\label{tab:main_results}
\begin{tabular}{lcccc}
\toprule
Method & $\auc_{130}$ & $\auc_{260}$ & Joint AUROC & TSR \\
\midrule
\ACal{} & 0.7004 & 0.7780 & $\mathbf{0.7392}\pm0.0480$ & $0.550\pm0.026$ \\
P-LINEAR & 0.6297 & 0.6575 & 0.6436 & 0.580 \\
P-TEMPORAL & 0.6268 & 0.6160 & 0.6214 & 0.580 \\
P-STATIC & 0.6433 & 0.5860 & 0.6146 & 0.580 \\
Frozen native & 0.4278 & 0.4693 & 0.4485 & 0.580 \\
\bottomrule
\end{tabular}
\end{table}

\paragraph{Confidence can be shaped.}
Frozen min-top confidence obtains joint AUROC 0.4485, and supervised frozen-feature probes obtain 0.6146--0.6436 (Table~\ref{tab:main_results}). \ACal{} reaches $0.7392\pm0.0480$. Using identical min-top summaries before and after adaptation increases AUROC from 0.4485 to 0.7327, ruling out summary choice as the sole explanation. It does \emph{not} hold visited states fixed: policy updates can alter the resulting trajectories.

\begin{figure}[h]
\centering
\includegraphics[width=0.90\linewidth]{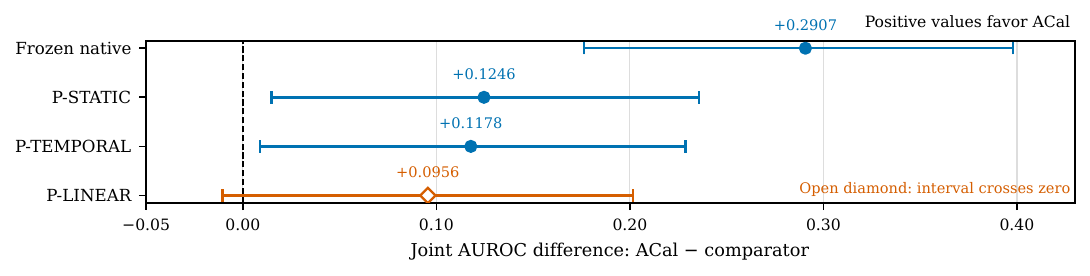}
\caption{Paired joint-AUROC differences with task-stratified 95\% bootstrap intervals. Positive values favor \ACal{}. The P-LINEAR interval includes zero, leaving its comparison unresolved.}
\label{fig:stage1_results}
\end{figure}

\paragraph{The strongest probe remains competitive.}
Paired intervals exclude zero for the gains over frozen native confidence, P-STATIC, and P-TEMPORAL (Figure~\ref{fig:stage1_results}). Against P-LINEAR, the point gain is 0.0956 but the interval is $[-0.0105,0.2017]$: the evidence does not establish superiority over this small readout. Its unchanged controller makes it an important practical comparator. Full intervals and seed results appear in Appendices~\ref{app:seed_results} and~\ref{app:uncertainty}.

\paragraph{Control performance must be considered.}
The adapted seeds succeed on 58, 54, and 53 of 100 test configurations versus 58 for the frozen controller, so mean TSR decreases from 0.58 to 0.55. LoRA changes may shift action-token rankings and thus selected actions; incorrect pair alignment or suppression of multiple valid actions are also plausible, \emph{untested} explanations. The three seeds do not establish a causal mechanism. Validation chose zero KL penalty, not evidence that preserving policy behavior is unnecessary. Appendix~\ref{app:mechanism} details these possibilities.

\paragraph{Scope and safety relevance.}
Results cover one checkpoint, one simulated benchmark, deterministic decoding, and three seeds. Normalized-time pairing can still confound physical task stages. Late-step AUROC uses a different active set, so its larger value is not evidence of better early warning. Real failure collection may be costly or hazardous (Appendix~\ref{app:deployment}); calibration, alarms, interventions, domain shift, and physical-robot safety are untested. Better ranking alone does not demonstrate a safer controller.

\section{Conclusion}
\CalVLA{} shapes native action confidence using task outcomes without an auxiliary detector. It improves failure discrimination relative to several frozen controls, but superiority over a linear readout is unresolved and mean task success falls. Confidence quality and controller performance must be evaluated jointly.

\clearpage
\bibliography{references}
\clearpage
\appendix

\section{Dataset census and split}
\label{app:split}

The evaluation uses the complete deterministic LIBERO-10 grid of ten tasks and 50 prescribed initial states per task. Frozen OpenVLA-7B succeeds on 243 of the 500 configurations and generates 202,906 observation--action steps. Initial states 0--9 enter training. Initial states 10--49 are permuted with seed 0 and divided into 25 additional training, five validation, and ten test states per task. The rule is outcome-blind and shared across tasks. Table~\ref{tab:split_appendix} gives the resulting composition.

\begin{table}[htbp]
\caption{Frozen configuration split and terminal outcomes.}
\label{tab:split_appendix}
\centering
\small
\begin{tabular}{lrrrr}
\toprule
Split & Configurations & Success & Failure & Success rate \\
\midrule
Train & 350 & 162 & 188 & 0.463 \\
Validation & 50 & 23 & 27 & 0.460 \\
Test & 100 & 58 & 42 & 0.580 \\
\midrule
Total & 500 & 243 & 257 & 0.486 \\
\bottomrule
\end{tabular}
\end{table}

Test outcomes remain sealed until selection is complete. No candidate is retrained, tuned, or selected using test performance.

\section{Pair construction}
\label{app:pairs}

For zero-indexed step $u\in\{0,\ldots,L-1\}$ in a rollout of length $L$, set $t=u+1$ and normalized elapsed time $\rho=t/L=(u+1)/L$, matching the one-indexed definition in Section~\ref{sec:method}. A successful and failed observation may be paired when they come from the same task and $|\rho^+-\rho^-|\leq0.02$. The canonical training artifact contains 11,378 pairs and 15,869 unique observations. Repeated observations are possible because a step can participate in more than one pair. The realized mean and maximum absolute progress differences are 0.000897 and 0.003239. Negative observations are drawn from the final 130 failed-trajectory steps. Each source episode contributes at most 128 pairs, and task-uniform sampling limits task imbalance.

The pair rows are not treated as independent robot trials. Statistical comparisons use held-out task--initial-state configurations as the resampling unit.

\section{Locked training configuration}
\label{app:training}

Each run makes one pass through a deterministic seed-specific permutation of the canonical pairs. Pair order is shared across $\lambda_{\mathrm{KL}}$ values within a seed. Table~\ref{tab:training_appendix} lists the recorded optimization settings. Confidence calculations, softmax, losses, LoRA parameters, and optimizer moments use FP32; the frozen backbone uses FP16.

\begin{table}[htbp]
\caption{Locked \ACal{} training configuration.}
\label{tab:training_appendix}
\centering
\small
\begin{tabular}{ll}
\toprule
Setting & Value \\
\midrule
Backbone & OpenVLA-7B, frozen FP16 \\
LoRA targets & $q_{\mathrm{proj}},v_{\mathrm{proj}}$ \\
Rank / scale & 16 / 16 \\
Dropout / bias & 0 / none \\
Trainable parameters & 8,388,608 \\
Optimizer & AdamW \\
Learning rate & $10^{-5}$ \\
Betas / epsilon & $(0.9,0.999)$ / $10^{-8}$ \\
Weight decay & 0 \\
Batch / accumulation & 1 pair / 1 \\
Updates / passes & 11,378 / 1 \\
Margin $\gamma$ / KL temperature & 0.2 / 1 \\
Scheduler / clipping & none / none \\
Early stopping / intermediate validation & disabled / disabled \\
\bottomrule
\end{tabular}
\end{table}

\section{Validation selection}
\label{app:selection}

The candidate grid contains six KL coefficients and three seeds, for 18 trained models. A family is eligible when mean validation TSR is at least 0.42 and every seed reaches at least 0.38. All six families pass. Eligible families are ranked by mean validation joint AUROC; differences below 0.01 are ties, resolved toward lower $\lambda_{\mathrm{KL}}$.

\begin{table}[htbp]
\caption{Validation results averaged over three seeds.}
\label{tab:family_validation}
\centering
\small
\begin{tabular}{rrr}
\toprule
$\lambda_{\mathrm{KL}}$ & Mean TSR & Mean joint AUROC \\
\midrule
\textbf{0} & 0.5733 & \textbf{0.6301} \\
1.5 & 0.5333 & 0.4761 \\
5 & 0.4800 & 0.5701 \\
15 & 0.5000 & 0.6297 \\
50 & 0.5467 & 0.5554 \\
150 & 0.5267 & 0.5655 \\
\bottomrule
\end{tabular}
\end{table}

The 0.0004 difference between $\lambda_{\mathrm{KL}}=0$ and 15 lies inside the 0.01 tie band, selecting 0. On validation, the selected family's difference from frozen native confidence is 0.0702 with paired interval $[-0.0808,0.2330]$. Validation chooses a family; it does not independently establish a resolved improvement.

\section{Frozen-probe validation}
\label{app:probes}

Frozen probes are selected without test outcomes. Their validation joint AUROCs are 0.7030 for P-LINEAR, 0.7318 for P-STATIC, and 0.7378 for P-TEMPORAL. P-STATIC and P-TEMPORAL are within the 0.01 tie band, so P-STATIC is the primary nonlinear comparator because it is temporally simpler. P-LINEAR remains mandatory as the least complex outcome-supervised readout. Its strongest test point estimate shows that comparator order is not stable across splits.

\section{Seed-level and comparator results}
\label{app:seed_results}

Average precision (AP) treats failure as the positive class. Table~\ref{tab:seed_appendix} reports every selected-family seed. Relative to frozen OpenVLA, mean $\Delta\tsr=0.550-0.580=-0.030$.

\begin{table}[htbp]
\caption{Locked test for the selected $\lambda_{\mathrm{KL}}=0$ family.}
\label{tab:seed_appendix}
\centering
\small
\setlength{\tabcolsep}{5pt}
\begin{tabular}{lrrrrrrr}
\toprule
Seed & Success & TSR & $\auc_{130}$ & $\auc_{260}$ & Joint & $\mathrm{AP}_{130}$ & $\mathrm{AP}_{260}$ \\
\midrule
0 & 58/100 & 0.580 & 0.6876 & 0.7325 & 0.7100 & 0.5980 & 0.7963 \\
1 & 54/100 & 0.540 & 0.6570 & 0.7690 & 0.7130 & 0.6399 & 0.8786 \\
2 & 53/100 & 0.530 & 0.7567 & 0.8324 & 0.7946 & 0.7529 & 0.9205 \\
\midrule
Mean & 55/100 & 0.550 & 0.7004 & 0.7780 & 0.7392 & 0.6636 & 0.8651 \\
SD & -- & 0.0265 & 0.0508 & 0.0501 & 0.0480 & 0.0801 & 0.0626 \\
\bottomrule
\end{tabular}
\end{table}

\begin{table}[htbp]
\caption{Locked frozen-policy comparators. All inherit frozen TSR 0.58.}
\label{tab:comparators_appendix}
\centering
\small
\begin{tabular}{lrrr}
\toprule
Comparator & $\auc_{130}$ & $\auc_{260}$ & Joint \\
\midrule
P-LINEAR & 0.6297 & 0.6575 & 0.6436 \\
P-TEMPORAL & 0.6268 & 0.6160 & 0.6214 \\
P-STATIC & 0.6433 & 0.5860 & 0.6146 \\
Frozen native min-top & 0.4278 & 0.4693 & 0.4485 \\
\bottomrule
\end{tabular}
\end{table}

\section{Interpretation of the confidence--control trade-off}
\label{app:mechanism}
The $\mathrm{AUROC}$ improvement does not establish that the policy has improved control or that confidence shaping caused the TSR change. In particular, \ACal{} changes the controller through the LoRA adapters: gradients that increase or decrease top-token probabilities can alter \emph{relative} action-bin scores and flip a greedy argmax. Multiplying all logits by the same positive scalar, by contrast, can change distribution sharpness without changing the argmax, so a confidence change does not necessarily imply an action change. Outcome labels also supervise entire rollouts; correct early actions in ultimately failed attempts may receive lower confidence, and increasing confidence in one successful action mode may discourage other valid actions. Normalized-time pairs may differ in their physical states and task stages. These are possible mechanisms, not causal conclusions of the present experiments. Isolating them would require analysis of changed action bins, physically matched trajectory pairs, task-stage labels, or policy-preservation objectives that maintain action quality. The current validation rule selects $\lambda_{\mathrm{KL}}=0$ based on its measured eligibility and discrimination, but this does not imply that KL-style preservation is generally undesirable.

\section{Practical data-collection considerations}
\label{app:deployment}
Outcome-supervised confidence shaping requires task-labeled successful and failed rollouts. In a physical robot, failed attempts can be expensive to collect and some failures can damage hardware or create hazards. A practical collection process would favor previously recorded operator-supervised attempts, simulator-generated failures checked against real observations, and conservative stop conditions during physical trials. These are deployment considerations rather than data-collection methods validated in this study. The supervised objective labels whole episodes; it does not reveal which specific action caused a failure. Adaptation may therefore alter useful action preferences, which is one possible explanation for the observed decrease in mean task success, but the experiments do not isolate that mechanism. Safe deployment would require tests of confidence calibration, alert thresholds, action-quality preservation, and intervention outcomes beyond the simulation results reported here.

\section{Paired uncertainty and score agreement}
\label{app:uncertainty}

Bootstrap resampling draws held-out configurations within each task and preserves method pairing. It does not resample training-pair rows. Table~\ref{tab:bootstrap_appendix} gives the point differences, bootstrap means, and 95\% intervals.

\begin{table}[htbp]
\caption{Paired task-stratified joint-AUROC comparisons.}
\label{tab:bootstrap_appendix}
\centering
\small
\begin{tabular}{lrrl}
\toprule
Comparator & Point diff. & Boot. mean & 95\% interval \\
\midrule
Frozen native & 0.2907 & 0.2888 & $[0.1764,0.3981]$ \\
P-STATIC & 0.1246 & 0.1242 & $[0.0148,0.2357]$ \\
P-TEMPORAL & 0.1178 & 0.1174 & $[0.0090,0.2287]$ \\
P-LINEAR & 0.0956 & 0.0951 & $[-0.0105,0.2017]$ \\
\bottomrule
\end{tabular}
\end{table}

Same-configuration Spearman correlations between \ACal{} and frozen-probe risks are descriptive rather than counterfactual: once the adapted policy takes a different action, later observations can diverge. The mean $\pm$ standard deviation across \ACal{} seeds is 0.2558$\pm$0.0371 at step 130 and 0.1705$\pm$0.1280 at step 260 for P-STATIC; 0.3248$\pm$0.0836 and 0.2324$\pm$0.1181 for P-LINEAR; and 0.3028$\pm$0.1021 and 0.2437$\pm$0.0961 for P-TEMPORAL.

\end{document}